# FrameNet Resource Grammar Library for GF

**Normunds Gruzitis, Peteris Paikens, Guntis Barzdins**

Institute of Mathematics and Computer Science, University of Latvia
Raina blvd. 29, Riga, LV-1459, Latvia
E-mail: {normunds.gruzitis, peteris.paikens, guntis.barzdins}@lumii.lv

**Abstract.** In this paper we present an ongoing research investigating the possibility and potential of integrating frame semantics, particularly FrameNet, in the Grammatical Framework (GF) application grammar development. An important component of GF is its Resource Grammar Library (RGL) that encapsulates the low-level linguistic knowledge about morphology and syntax of currently more than 20 languages facilitating rapid development of multilingual applications. In the ideal case, porting a GF application grammar to a new language would only require introducing the domain lexicon – translation equivalents that are interlinked via common abstract terms. While it is possible for a highly restricted CNL, developing and porting a less restricted CNL requires above average linguistic knowledge about the particular language, and above average GF experience. Specifying a lexicon is mostly straightforward in the case of nouns (incl. multi-word units), however, verbs are the most complex category (in terms of both inflectional paradigms and argument structure), and adding them to a GF application grammar is not a straightforward task. In this paper we are focusing on verbs, investigating the possibility of creating a multilingual FrameNet-based GF library. We propose an extension to the current RGL, allowing GF application developers to define clauses on the semantic level, thus leaving the language-specific syntactic mapping to this extension. We demonstrate our approach by reengineering the MOLTO Phrasebook application grammar.

## 1 Introduction

Controlled natural languages (CNL) can be divided into two general types according to the formalist or the naturalist approach [1]. The formalist approach supports a deterministic, bidirectional mapping of CNL to a formal language like first-order logic (FOL) or, more commonly, to description logic, namely OWL (Web Ontology Language) [2], allowing the integration with existing tools for reasoning, consistency checking and model building. Although logic-based CNL provides a seemingly informal high-level means for knowledge representation, essentially it is still a formal language that is just as expressive as the corresponding formalism, and whose interpretation is deterministic (predictable). In contrast, in the naturalist approach possible ambiguities are decreased but not excluded, thus allowing for a wider coverage of NL and more informal applications, such as semantically precise machine translation within a CNL.

In other words, there are CNLs that have an underlying logic-based formalism defining the semantics of a text (e.g. Attempto Controlled English [3]), and there are CNLs that do not have an underlying logic-based formalism (e.g. MOLTO Phrasebook [4] for multilingual translation of touristic phrases). In the first case, the semantics of CNL statements is interpreted by both a human and a formal-logic reasoning machine. In the second case, the interpreter is primarily a human and possibly a domain-specific application that uses CNL, for example, for information retrieval from a predefined domain-specific database.

Grammatical Framework (GF) [5] is a categorial grammar formalism and a toolkit for programming multilingual grammar applications. It is similar to definite clause grammars (DCG) in Prolog in that both support parsing and synthesis using the same (categorial) grammar definition. Besides the grammar formalism itself, an important part of GF is its Resource Grammar Library (RGL) [6] that encapsulates the low-level linguistic knowledge about morphology and syntax of currently more than 20 languages (the number is constantly growing). RGL facilitates rapid development and porting of application grammars in many parallel languages: all GF resource grammars implement the same syntactic interlingua (API) enabling automatic translation among languages via the abstract syntax trees. In particular, it has been shown that GF is a convenient framework for rapid and flexible implementation of multilingual CNLs – both those rooted in a formal language like the FOL-based Attempto Controlled English [7] and those rooted in a relatively informal language like standard touristic phrases [4].

In the ideal case, porting a GF application grammar to a new language or domain would only require introducing the domain lexicon – translation equivalents that are interlinked via common abstract terms. While it is possible for a highly restricted CNL, e.g. for authoring and verbalizing OWL ontologies (as implemented, for example, in [8]), developing and porting a less restricted CNL requires more linguistic knowledge about the particular language, and more experience in GF (particularly, in using RGL). Specifying a lexicon of nouns (incl. multi-word units) is mostly straightforward, however, specifying the lexicon of verbs is typically the most complex task in terms of both inflectional paradigms and argument structure, and may require specifying the whole clause (as in Phrasebook). Thus adding verbs to a GF application grammar's lexicon in a foreign language (or for a novice or less resourced GF developer even in his mother tongue) might not be a straightforward task and might present a stumbling block for potential GF multilingual application developers. Therefore in this paper we are focusing on verbs, investigating the possibility of creating a multilingual FrameNet-based GF resource grammar library.

The rest of the paper is organized as follows. In Section 2 we briefly re-capture the relevant architectural principles of FrameNet. In Section 3 we similarly re-capture some relevant GF application grammar development principles, demonstrating the current approach with a detailed example. Section 4 modifies the previous example, describing our solution for integrating FrameNet into GF application grammars.

Finally we conclude with a brief discussion on the potential and benefits of the proposed FrameNet library for seamless multilingual CNL development.

## 2 FrameNet

FrameNet [9] is a lexicographic database that describes word meanings based on the principles of frame semantics. The central idea of frame semantics is that word meanings must be described in relation to semantic frames [10]. Therefore, the *frame* and the *lexical unit* are the key components of FrameNet. A lexical unit in FrameNet terms is the combination of a lemma with a specific meaning – each separate meaning of a word represents a new lexical unit. In FrameNet, each lexical unit is related to a semantic frame that it is said to *evoke* a frame (see Figure 1 and Figure 2).

The semantic frame describes a certain situation and the participants of that situation that are likely to be mentioned in the sentences where the evoking lexical unit (referred to as frame *target*) appears. The semantic roles played by these participating entities are called *frame elements* (FE). FrameNet makes a differentiation between *core* frame elements and *peripheral* frame elements. In general, frame elements that are necessarily realized are core elements. Peripheral elements represent more general information such as time, manner, place, and purpose and are less specific to the frame. Nevertheless all FrameNet frame elements are local to individual frames. This avoids the commitment to a small set of universal roles, whose specification has turned out to be controversial in the past [11]. In order to account for actual similarities between frame elements in different frames FrameNet includes also a rich set of frame to frame and FE to FE relations.

|  | `Residence` | *This frame has to do with people (the Residents) residing in Locations, sometimes with a Co-resident.* |
|---|---|---|
| Core FEs | `Co_resident` | *A person or group of people that the Resident is staying with or among.* |
| | `Location` | *The place in which somebody resides.* |
| | `Resident` | *The individual(s) that reside at the Location.* |
| | Lexical units | camp.v, dwell.v, inhabit.v, live.v, lodge.v, occupy.v, reside.v, room.v, squat.v, stay.v |

**Fig. 1.** A sample FrameNet frame (only core frame elements shown).

The frame descriptions are coarse-grained and generalize over lexical variation. Therefore lexeme-specific information is contained within lexical unit entries that are more fine-grained and contain a definition of the lexical unit, the syntactic realizations of each frame element and the valence patterns. A sense of a lemma (word meaning) can evoke a frame, and thus form a lexical unit for this frame, if this sense is syntactically able to realize the core frame elements that instantiate a conceptually necessary component of a frame [12].

In Figure 2, a simplified (summarized) lexical entry of 'to live' (`Residence`) is given: information on non-core FEs is excluded (the rest is summed up); for each valence model only the most frequent realization pattern is given; valence models that contain multiple FEs of the same type are excluded; valence models that have appeared in the corpus only once are excluded; prepositional phrase patterns (`PP`) are not distinguished by particular prepositions.

| FE | Total | Pattern |
|---|---|---|
| `Co_resident` | 14 | PP.Dep (86%) |
| `Location` | 131 | PP.Dep (81%) |
| `Resident` | 143 | NP.Ext (90%) |

(a)

| Total | Patterns | | |
|---|---|---|---|
| | | Location | Resident |
| 98 | | | |
| 71% | | PP.Dep | NP.Ext |
| 7 | `Co_resident` | | `Resident` |
| 86% | PP.Dep | | NP.Ext |
| 7 | `Co_resident` | `Location` | `Resident` |
| 86% | PP.Dep | PP.Dep | NP.Ext |

(b)

**Fig. 2.** A simplified lexical entry `Residence.live`. (a) Core FEs and their most frequent syntactic patterns in the FrameNet corpus. (b) Most frequent valence models of core FEs.

Our central point of interest in this paper is the multilingual dimension of FrameNet. A number of projects have investigated the use of English FrameNet frames for other languages, such as German (SALSA project [13]), Spanish [14], Japanese [15], and lately also for Thai, Chinese, Italian, French, Bulgarian, Hebrew [16]. A fundamental assumption of these projects is that English FrameNet frames can be largely re-used for the semantic analysis of other languages. This assumption rests on the nature of frames as coarse-grained semantic classes which refer to prototypical situations – to the extent that these situations agree across languages, frames should be applicable cross-linguistically. Also Boas [17] suggests the use of semantic frames as interlingual representation for multilingual lexicons.

While FrameNet multilinguality is clearly a very attractive assumption, its empirical validation comes primarily from the German SALSA project, which has found that the vast majority of English FrameNet frames can be directly applied to the analysis of German – a language that is typologically close to English. Meanwhile, some frames have turned out not to be fully interlingual and three main cross-lingual divergence types were found:

1. Ontological distinctions between similar frame elements.
2. Missing frame elements.
3. Differences in lexical realization patterns (e.g. German 'fahren' does not distinguish between `Operate_vehicle` and `Ride_vehicle` frames).

Nevertheless this empirical evidence shows that most of FrameNet frames indeed are language independent and therefore provide an opportunity for use as a multilingual coarse-grained lexicon in GF, as described in Section 4. Although FrameNet addresses all parts-of-speech, its strength and focus is on verbs for which the best coverage is provided. This is largely because while noun-phrase multi-word units are extensively "invented" to denote nominal concepts (especially in technical domains), phrasal verbs are more fixed, commonly reused (across domains) and are often captured in dictionaries of standard language. Since the advantage of valence structures is more obvious for verbs, in this paper we consider only FrameNet frames with verbal frame evoking lexical entries.

## 3  GF Application Grammar Development: the Current Approach

GF facilitates reusability by splitting the grammar development in two levels:

1. A general-purpose *resource grammar* covers a wide range of morphological paradigms and syntactic structures and as such is highly ambiguous. GF provides a Resource Grammar Library (RGL) [6] implementing a common API for more than 20 languages.
2. Domain specific *application grammars* reuse the RGL, defining semantic structures and the subset of natural language (syntax and lexicon) that is used within a particular CNL. Application grammars reduce or even eliminate ambiguities.

Development of a resource grammar requires in-depth GF knowledge and in-depth linguistic knowledge about the particular language. Once a resource grammar is provided, application grammars are built on top of it significantly reducing the linguistic knowledge prerequisites (a non-linguist native or fluent speaker should be sufficient), as well as he or she can be less experienced with GF.

GF differentiates not only between general-purpose resource grammars and domain-specific application grammars, but also between *abstract syntax* and *concrete syntax*. The abstract syntax captures the semantically relevant structure of a CNL, defining grammatical categories and functions for building abstract syntax trees [5]. The concrete syntax defines the linearization of the CNL abstract syntax trees at the surface level for each language. Translation among languages (concrete syntaxes) is provided via abstract syntax[1].

We will describe the current approach to the RGL-based GF application grammar development using the MOLTO Phrasebook application [4] for multilingual translation of touristic phrases as an example. Phrasebook is a CNL implemented in 15 languages and is aimed to be usable by anyone without prior training. It has 42 categories and 290 functions. The number of phrases it can generate is infinite, but on a reasonable level of tree depth 3, Phrasebook has nearly 500,000 abstract syntax trees [4]. We will consider only a small subset of the Phrasebook grammar – categories (`cat`) and functions or constructors (`fun`) that are used to build the abstract syntax trees for the following sample sentences, and to generate these sentences from the corresponding abstract trees[2] as given in Figure 3.

---

[1] Note that in GF there is no concept of a language pair or a translation direction. Also there is no common semantic interlingua. Instead there are many application specific (i.e., CNL and domain specific) interlinguas.

[2] The provided abstract syntax trees are slightly simplified regarding the pronouns – their gender, number and politeness features – to avoid multiple variants.

In the next section we will modify the English implementation of the Phrasebook grammar by means of the proposed FrameNet-based resource grammar, acquiring a simpler English Phrasebook (PhrasebookEng) implementation as the result, while preserving the same functionality. However, we will not make any changes neither in the Phrasebook functor (the common incomplete concrete syntax), nor in the abstract syntax, i.e., we will not impose any special requirements on application grammar design.

| English sentences | Phrasebook abstract syntax |
|---|---|
| *I like this pizza.* | `PSentence (SProp (PropAction (ALike I (This Pizza))))` |
| *I live in Belgium.* | `PSentence (SProp (PropAction (ALive I Belgium)))` |
| *I love you.* | `PSentence (SProp (PropAction (ALove I You)))` |
| *I want a good pizza.* | `PSentence (SProp (PropAction (AWant I (OneObj`<br>`   (ObjIndef (SuchKind (PropQuality Good) Pizza))))))` |
| *I want to go to a museum.* | `PSentence (SProp (PropAction (AWantGo I`<br>`   (APlace Museum))))` |

**Fig. 3.** Sample Phrasebook sentences along with their abstract syntax trees.

The abstract syntax of Phrasebook that represents the syntactic and semantic model of the above phrases is given in Figure 4.

```
cat
  Action ;     -- proposition about a Person, e.g. "I love you"
  Phrase ;     -- complete phrase, e.g. "I love you."
  Country ;    -- e.g. "Belgium"
  Item ;       -- single entity, e.g. "this pizza"
  Kind ;       -- kind of an item, e.g. "pizza"
  Object ;     -- e.g. "a good pizza"
  Person ;     -- agent wanting or doing something, e.g. "I"
  Place ;      -- location, e.g. "a museum"
  PlaceKind ;  -- kind of location, e.g. "museum"
  Property ;   -- basic property of an item, e.g. "good"

fun
  Belgium : Country ;
  Good    : Property ;
  Museum  : PlaceKind ;
  Pizza   : Kind ;

  ALike   : Person -> Item    -> Action ;  -- Action(Person, Item)
  ALive   : Person -> Country -> Action ;
  ALove   : Person -> Person  -> Action ;
  AWant   : Person -> Object  -> Action ;
  AWantGo : Person -> Place   -> Action ;
```

**Fig. 4.** A fragment of Phrasebook abstract syntax (semantic model).

A fragment of the incomplete concrete syntax (aka functor) of Phrasebook is given in Figure 5. This is a technical intermediate layer between the abstract syntax and its implementation in concrete syntaxes. It defines language-independent syntactic categories and structures (e.g. `PSentence`, `SProp`, `PropAction`) that are common to all (or most) languages. Thus the concrete syntax of a particular language has to specify only language-dependent structures and the lexicon (e.g. `AWant`, `Good`, `Pizza`).

Note that the functor defines the mapping between the application-specific abstract syntax categories and the categories of the Resource Grammar Library. For instance, `Country`, `Item` and `Object` syntactically are realized as noun phrases (category `NP` in RGL). In Figure 5, there are also three Phrasebook categories that are not directly mapped to RGL categories (`Person`, `Place` and `PlaceKind`). Instead, they are defined as application-specific categories `NPPerson`, `NPPlace` and `CNPlace` that are specified as record types whose fields (e.g. `name`, `at`, `to`) are of RGL types.

```
lincat   -- category linearization types
  Phrase = Text ;
  Action = Cl ;
  Country, Item, Object = NP ;
  Person = NPPerson ;
  Place = NPPlace ;
  Kind = CN ;
  PlaceKind = CNPlace ;
  Property = A ;

oper   -- operations - functions in concrete syntax
  NPPerson : Type = {name : NP ; isPron : Bool ; poss : Quant} ;
  NPPlace  : Type = {name : NP ; pos : Adv ; dir : Adv} ;
  CNPlace  : Type = {name : CN ; pos : Prep ; dir : Prep} ;

  mkNPPerson : Pron -> NPPerson = \pron ->
    {name = mkNP pron ; isPron = True ; poss = mkQuant pron} ;

  mkCNPlace : CN -> Prep -> Prep -> CNPlace = \cn,prep1,prep2 ->
    {name = cn ; pos = prep1 ; dir = prep2} ;

  mkNPPlace : Det -> CNPlace -> NPPlace = \det,place ->
    let name : NP = mkNP det place.name in {
      name = name ;
      pos = mkAdv place.pos name ;   -- place - position
      dir = mkAdv place.dir name     -- place - direction
    } ;
```

**Fig. 5.** A fragment of Phrasebook incomplete concrete syntax (functor): common structures. To make the code more intelligible to readers unfamiliar with GF, it has been slightly modified.

The predication patterns (`Action`) with verbs at the centre are perhaps the most complex functions in Phrasebook (from the implementation point of view). Note that these actions are of type `Cl` (clause): this will be the gluing point for the integration of the FrameNet-based resource library (see Section 4). The linguistic (English) realization of the semantic model is specified by the concrete syntax (given in Figure 6), which tells how abstract syntax trees are linearized (`lin`) into English strings. The same rules are also used for parsing.

```
lin  -- function linearization rules
  Belgium = mkNP (mkPN "Belgium") ;
  Good = LexiconEng.good_A ;
  Museum = mkPlaceKind "museum" "at" ;
  Pizza = mkCN (mkN "pizza") ;

  ALike pers item = mkCl pers.name (mkV2 (mkV "like")) item ;
  ALive pers country = mkCl pers.name (mkVP (mkVP (mkV "live"))
    (mkAdv SyntaxEng.in_Prep country)) ;
  ALove pers1 pers2 =
    mkCl pers1.name (mkV2 (mkV "love")) pers2.name ;
  AWant pers obj = mkCl pers.name (mkV2 (mkV "want")) obj ;
  AWantGo pers place = mkCl pers.name SyntaxEng.want_VV
    (mkVP (mkVP IrregEng.go_V) place.dir) ;
oper
  mkPlaceKind : Str -> Str -> CNPlace = \name,prep_pos ->
    mkCNPlace (mkCN (mkN name)) (mkPrep prep_pos) SyntaxEng.to_Prep ;
```

**Fig. 6.** A fragment of Phrasebook concrete syntax for English.

Verbs, in general, are at the centre of a sentence, both syntactically and semantically. They have the most complex inflectional paradigms (at least in inflective languages). The syntactic and semantic valence of a verb is defined via its argument and modifier structure. This inevitably requires solid linguistic knowledge.

RGL differentiates among V (intransitive), V2 (transitive) and V3 (ditransitive) verbs, as well as some more specific types of verbs with syntactically fixed argument structure. The syntactic valence patterns for the predefined verb types are fixed when defining a verb in the application lexicon; these patterns do not depend on the argument (i.e. the case or preposition of the argument does not depend on a particular NP). Other valences are specified while constructing a verb phrase – as adverbial modifiers (`Adv`); their syntactic patterns are specified by the application developer for each target language, depending on the semantic role of the argument and syntactic properties of the language.

If compared to nouns, there are much less verbs and they are more ambiguous (see WordNet statistics[3], for example), thus verbs are also more reusable linguistic units than nouns. This suggests that a reusable lexicon of verbs would be helpful for GF application developers. However, this requires not only a dictionary, but also additional information about the basic syntactic valences for the direct and indirect objects. Even more helpful would be a multilingual resource grammar of verb valences.

---
[3] http://wordnet.princeton.edu/wordnet/man/wnstats.7WN.html

There is a small, limited multilingual lexicon provided by the RGL, but it does not provide systematic means for scaling and expanding beyond the V-, V2- and V3-like valences. The basic lexicon also does not support polysemous verbs – valences often are different for various meanings of the same verb, and vice versa.

An impression of the syntactic coverage of the GF Resource Grammar Library can be obtained from its API documentation[4]. Figure 7 illustrates some of the constructors for clauses, verb phrases, noun phrases, common nouns, and adverbial modifiers that are referred in Figure 5 and Figure 6. For instance, Figure 7 illustrates that a clause can be built from a subject noun phrase with a verb and appropriate arguments. In general, a clause can be built from a subject noun phrase and a verb phrase.

| Function | Type | Example |
|---|---|---|
| `mkCl` | `NP -> VP -> Cl` | *she always sleeps* |
| `mkCl` | `NP -> V2 -> NP -> Cl` | *she loves him* |
| `mkCl` | `NP -> VV -> VP -> Cl` | *she wants to sleep* |
| `mkVP` | `VP -> Adv -> VP` | *to sleep here* |
| `mkNP` | `Det -> CN -> NP` | *the old man* |
| `mkNP` | `PN -> NP` | *Paris* |
| `mkNP` | `Pron -> NP` | *we* |
| `mkCN` | `N -> CN` | *house* |
| `mkAdv` | `Prep -> NP -> Adv` | *in the house* |

**Fig. 7.** A fragment of the Resource Grammar API documentation.

In order to use the proposed FrameNet library (in addition to the Resource Grammar API), the application grammar developer will have to consult the FrameNet API as presented in the next section.

## 4 Our FrameNet-Based Approach

The current split of functionality between the application and the common libraries expects applications to define the domain specific knowledge in all languages by using specific verbs and defining their syntactic valences in each target language.

Our proposal is to raise the abstraction level for the common GF clause construction API from the current syntactic definition to a more semantic one. As we discussed in Section 2, the research on frame semantics suggests that an exhaustive cross-domain linguistic model of semantic frames and roles is possible, and it has been implemented for multiple languages. We believe that it is possible and reasonable to facilitate the development of multilingual application grammars in GF by referencing a common API of semantic frames that provide language-specific linearization for whole clauses or verb phrases (VP), and can optionally provide a default choice of a lexical unit that evokes the frame and default syntactic valence patterns.

---
[4] http://www.grammaticalframework.org/lib/doc/synopsis.html

In particular, we envision a resource grammar library that is built on top of the current RGL offering each of the FrameNet's semantic frames as a function that builds a clause from given parameters: fillers of the core elements of that frame, and an optional list of elements filling the peripheral roles. The FrameNet API would be implemented semi-automatically by generating GF code from FrameNet data providing a set of overloaded functions for each frame – mapping the frame (its elements) to the default or specific syntactic realization (linearization). Our observation is that, in the current approach to GF application development, a miniature ad-hoc 'framenet' is actually implemented for each application. Moreover, it is often 'reused' in a copy-paste-edit manner from previous applications or from concrete syntaxes of other languages that implement the same application. We would like to promote systematic means for reusing this language-specific knowledge via common language-independent frames.

The following simplified assumptions underlie the default behaviour of our approach (the default behaviour can be overridden for specific syntactic patterns and lexical units – see systematic "exceptions" illustrated below):

1. For each frame element there is a typical syntactic pattern that is used in most cases – independently of the verb that evokes the frame. I.e., both semantic and syntactic valences can be defined at the frame level.
    1.1. There is a common syntactic realization of a frame (a clause or a verb phrase) that is reused by most verbs that evoke the frame.
2. It is possible to specify a default lexical unit (the most general and/or the most frequently used verb) that evokes the frame, so that it can be used in the linearization (translation) of the frame, if no specific verb is provided.
3. In the CNL settings, it is often sufficient that only core semantic valences (core frame elements according to FrameNet) are available.

These assumptions, of course, do not hold in general, but they help us to keep the presentation of our approach simpler. Even then we cannot fully isolate the application developer from providing some language-specific features. For example, the Phrasebook application in English (and similarly in Russian) needs to distinguish between locations that are "at place" or "in place" – the preposition does not depend on the frame and not even on the specific verb, but on the particular noun (the filler of a frame element). In contrast to the highly analytical English, in many languages it might be necessary to customize the realization of the whole clause, depending on the verb. For example, in Latvian (and similarly in Russian, Italian and German) there are verbs (systematic "exceptions") that instead of the subject in the nominative case and the object in the accusative case require the subject in the dative case and the object in the nominative case[5] (see Figure 8). In the actual implementation of the FrameNet RGL, such agreement variations have to be handled by alternative verb-specific clauses implemented in the frame functions.

---

[5] Here we use the term 'case' in a broad sense: in Italian, for example, there are no cases for nouns; cases are expressed by prepositions, pronouns and implicitly by word order.

|  | **LOVE** | **LIKE** |
|---|---|---|
| English | I$_{[Nom]}$ love pizza$_{[Acc]}$ | I$_{[Nom]}$ like pizza$_{[Acc]}$ |
| German | Ich$_{[Nom]}$ liebe Pizza$_{[Acc]}$ | Ich$_{[Nom]}$ mag Pizza$_{[Acc]}$<br>Mir$_{[Dat]}$ gefällt Pizza$_{[Nom]}$ |
| Italian | Io$_{[Nom]}$ amo la pizza$_{[Acc]}$ | A me$_{[Dat]}$ piace la pizza$_{[Nom]}$ |
| Latvian | Es$_{[Nom]}$ mīlu picu$_{[Acc]}$ | Man$_{[Dat]}$ patīk pica$_{[Nom]}$ |
| Russian | Я$_{[Nom]}$ люблю пиццу$_{[Acc]}$ | Мне$_{[Dat]}$ нравится пицца$_{[Nom]}$ |

**Fig. 8.** Verb-specific realization of the frame elements Experiencer and Content in different languages. All these verbs belong to the Experiencer_focus frame.

As in the case of the syntactic RGL, the proposed semantic resource grammars of the FrameNet library will also be ambiguous as such: the same verb can evoke different frames, and the same frame might be evoked by contradicting verbs (e.g. both 'to love' and 'to hate' evoke the same Experiencer_focus frame). However, the intuition is that the developer of a domain-specific CNL will reduce or eliminate the semantic ambiguity by avoiding ambiguous mappings between lexical units and frames, by specifying concrete verb lexemes instead of relying on the default ones etc. – analogically as it is currently done at the syntactic level.

To illustrate the use of the FrameNet API, we provide a sample re-implementation of some clause-building functions from the MOLTO Phrasebook application (see Figure 9) in contrast to the current PhrasebookEng implementation of the same functions as shown earlier in Figure 6.

```
Before:
ALike pers item = mkCl pers.name (mkV2 (mkV "like")) item ;
ALive pers country = mkCl pers.name (mkVP (mkVP (mkV "live"))
  (mkAdv SyntaxEng.in_Prep country)) ;
ALove pers1 pers2 =
  mkCl pers1.name (mkV2 (mkV "love")) pers2.name ;
AWant pers obj = mkCl pers.name (mkV2 (mkV "want")) obj ;
AWantGo pers place = mkCl pers.name SyntaxEng.want_VV
  (mkVP (mkVP IrregEng.go_V) place.dir) ;
After:
ALike pers item =
  Experiencer_focus (mkV "like") pers.name item NIL NIL ;
ALive pers country = Residence pers.name NIL country ;
ALove pers1 pers2 =
  Experiencer_focus (mkV "love") pers1.name pers2.name NIL NIL ;
AWant pers obj = Possession (mkV "want") pers.name obj ;
AWantGo pers place =
  Desiring pers.name (Motion_VP IrregEng.go_V NIL place.name) ;
```

**Fig. 9.** Changes to the PhrasebookEng syntax using the proposed FrameNet API.

As seen in Figure 9, the application grammar developer still has to provide the domain-specific knowledge that the application requires, and some simple constructors of the GF RGL are still used, but the code is more intelligible and 'flat' – it is not specified how the parameters (frame elements) are glued together to build up verb phrases and clauses[6]. The proposed API refers to the semantic roles only: if the user specifies, for example, the resident of the `Residence` frame (`ALive` action in Phrasebook), the FrameNet library maps it to the relevant syntactic role (subject in this case). Thus the verb and clause building part of application grammars such as Phrasebook is in essence reduced to mapping domain-specific concepts to the appropriate general FrameNet frames, and to specifying the omitted core frame elements, if any (`NIL`[7]).

In multilingual applications, there is a general issue of selecting lexical units – translation equivalents. For example, for the `Residence` frame, there are many possible verbs that describe the same situation with various semantic nuances (e.g. 'to camp', 'to dwell', 'to live', 'to stay'; see Figure 1). If these differences are relevant to the application domain, then a particular lexical unit can be explicitly specified. If the differences are not considered important for a particular use-case or concept, the preferred lexical unit for the chosen frame can be omitted, resulting in a robust system that would use a default verb (e.g. 'to live') when generating a text, and that would allow all frame-relevant verbs in parsing.

An advantage of this approach is the ability to build robust multilingual CNL applications without expertise in all covered languages. The benefit of using GF is that it would be possible to port such applications to other languages without going into details of their grammars – as they are already implemented in the common RGL. Furthermore, it is possible to omit the details about how the semantic roles are mapped to syntactic elements, as the same semantic element may be expressed by different syntactic means when translating the same clause to another language.

The API of the proposed FrameNet RGL is illustrated in Figure 10 (similarly as the API of GF RGL in Figure 7). The function names match the FrameNet frame names, thus the API can be automatically documented by FrameNet data providing definitions and examples for each frame (function) and each frame element (argument of the function).

Although currently we have handcrafted the code of the sample FrameNet library, we have done it systematically using the actual FrameNet data that is well structured and includes statistics from a FrameNet-annotated corpus. This has given confidence that FrameNet data can be used to automatically generate both the abstract syntax of the FrameNet API and its implementation for English and other languages using the current GF RGL syntactic categories and constructors, and properly addressing verb-specific valence patterns.

---

[6] Note that the implementation of the `AWantGo` function is not 'flat' – there are nested frames. I.e., it might be necessary to specify the semantic tree structure, but not the syntactic structure.

[7] We have not specified the implementation of `NIL` arguments yet, but this is only a technical mater.

We have performed some initial experiments on automatic GF code generation from FrameNet data, but the development of a more elaborated convertor is pending. Nevertheless, there are only about 1000 frames in FrameNet, therefore the generated code can also be manually debugged and improved afterwards.

| Function/Frame | Type | Mapping to FEs |
|---|---|---|
| Residence | `V -> NP -> PP -> Adv -> Cl` | Resident → Co_resident → Location |
| | `V -> NP -> NIL -> Adv -> Cl` | |
| | `NP -> NIL -> NP -> Cl` | |
| Possession | `V -> NP -> NP -> Cl` | Owner → Possession |
| | `NP -> NP -> Cl` | |
| Desiring | `VV -> NP -> VP -> Cl` | Experiencer → Event |
| | `NP -> VP -> Cl` | |
| Motion | `V -> NP -> NP -> NP -> Cl` | Theme → Source → Goal |
| | `NP -> NP -> NP -> Cl` | |
| Motion_VP | `V -> NP -> NP -> VP` | Source → Goal |
| | `V -> NIL -> NP -> VP` | |
| | `NP -> NP -> VP` | |
| | `Adv -> Adv -> VP` | |
| Experiencer_focus | `V -> NP -> NP -> VP -> NP -> Cl` | Experiencer → Content → Event → Topic |
| | `V -> NP -> NP -> NIL -> NIL -> Cl` | |
| | `NP -> NP -> NIL -> NIL -> Cl` | |

**Fig. 10.** A simplified fragment of the proposed FrameNet API. The `Desiring` frame has actually four core elements, and `Motion` – seven. Also all the possible combinations of `NP`, `PP`, `Adv` and `NIL` argument types are not included. Note that the `Motion_VP` is a special case of `Motion` – generated for use as a nested frame (as the `VP` object of a `VV` verb).

The manually generated code for several FrameNet frames, as shown in Figure 11, implements the features in a very similar manner as the Phrasebook application shown earlier in Figure 6 – which is to be expected, as it needs to realize similar syntactic structures with the same GF resources. However, a major difference is that this code would be reusable for multiple applications, and it could cover larger domains in a scalable way.

There are still some technical issues that need to be addressed, such as a more convenient way for specifying omitted core frame elements, but we believe that these are minor challenges. A particular concern is common peripheral semantic roles such as Time, Place and Manner that are encountered in nearly all frames (if they are not among the core roles for that frame). Again, the current Resource Grammar API deals with them on a syntactic level – providing means to attach various adverbial modifiers. We propose adding them as a (possibly empty) list of peripheral parameters, allowing the language-specific API implementation to handle the word order changes as needed.

```
-- Residence : NP -> NIL -> NP -> Cl
Residence resident NIL location = Residence (mkV "live") resident NIL
  (mkAdv SyntaxEng.in_Prep location) ;

-- Residence : V -> NP -> NIL -> Adv -> Cl
Residence verb resident NIL location =
  mkCl resident (mkVP (mkV "live") location) ;

-- Residence : V -> NP -> PP -> Adv -> Cl
Residence verb resident co_resident location = mkCl resident
  (mkVP (mkVP (mkV2 verb co_resident.prep) co_resident.np) location) ;

-- Possession : V -> NP -> NP -> Cl
Possession verb owner possession =
  mkCl owner (mkVP (mkV2 verb) possession) ;

-- Desiring : NP -> VP -> Cl
Desiring experiencer event =
  Desiring SyntaxEng.want_VV experiencer event ;

-- Desiring : VV -> NP -> VP -> Cl
Desiring verb experiencer event = mkCl experiencer verb event ;

-- Motion : V -> NP -> NP -> NP -> Cl
Motion verb theme source goal = mkCl theme (Motion_VP verb source goal) ;

-- Motion_VP : NP -> NP -> VP
Motion_VP source goal = Motion_VP (mkV "move") source goal ;

-- Motion_VP : V -> NP -> NP -> VP
Motion_VP verb source goal = mkVP (
  (mkVP (mkVP verb) (mkAdv SyntaxEng.from_Prep source))
  (mkAdv SyntaxEng.to_Prep goal)) ;

-- Motion_VP : V -> NIL -> NP -> VP
Motion_VP verb NIL goal =
  mkVP (mkVP verb) (mkAdv SyntaxEng.to_Prep goal) ;

-- Experiencer_focus : V -> NP -> NP -> NIL -> NIL -> Cl
Experiencer_focus verb experiencer content NIL NIL =
  mkCl experiencer (mkV2 verb) content ;
```

**Fig. 11.** English implementation of the proposed FrameNet API (a simplified fragment). `PP` extends the RGL set of categories; its linearization type is `{prep : Prep ; np : NP}`.

## 5  Discussion and Future Work

Currently the GF toolset provides a reusable syntactic framework for the development of multilingual domain-specific CNLs. When one acquires a solid understanding of the RGL structure and design principles, and gets used to the RGL-based application grammar design patterns, it is a rather rapid development to provide a concrete syntax for a language he or she knows well[8]. However, the process still might not be straightforward, especially when porting a third-party application, as it might not be enough to look at the code of e.g. English implementation to (immediately) understand the intended meaning of a specific abstract word or clause to provide an appropriate translation. In addition, different application grammars that cover related domains will more or less overlap, so that the same structures are re-implemented for each application.

In the current approach, GF application grammar developers essentially provide a miniature domain-specific framenet for each application. We make a case for basing application development on a common, reusable semantic framework, and argue that it is reasonably possible to develop such a framework by leveraging the existing FrameNet data. Working on the semantic level requires specific knowledge and training as well[9], but the resulting systems are more generic and easier to reuse across languages and across applications and domains.

The proposed approach is aimed to lower the entrance barrier of the GF application grammar development by moving it from the language-specific syntactic level towards the language-independent semantic level. The long-term goal is to facilitate the development of multilingual applications by providing robust means for automatic alignment of translation equivalents (particularly verbs), and by reducing syntactic and lexical ambiguities that appear in the parsing and generation of less restricted CNLs. GF has been chosen as an advanced and well-resourced framework for this purpose, but the proposed general principle could be applied also to other grammar formalisms.

The main limitations of the proposed approach to some extent are related to the limitations of FrameNet, particularly its coverage (in terms of lexical units). Furthermore, the coverage might differ among languages. The list of the lexical units for each frame could be extended via WordNet, as it has been shown by Johansson and Nugues [18], however then we would have to fall back to the frame-specific (vs. verb-specific) valence patterns. Another limitation is that even the most frequently used verb-specific valence patterns might not be appropriate in specific cases.

Although we have tested our proposal only on the English FrameNet data and English Phrasebook grammar, considering other languages only theoretically, we believe that in overall this would ease the multilingual GF application development, and that the limitations can be overcome by using the RGL syntactic structures directly where necessary. By relying on the default syntactic realization and the default lexical units, one can quickly obtain the first working version of a multilingual aplication for further testing and tuning. In fact, the default behaviour can be specified already in the functor that defines the common language-independent structures of an application grammar.

---

[8] The experience of the MOLTO team shows that adding a new language to Phrasebook takes 1.5 days on average.

[9] GF developers would have to explore and consult the documentation of FrameNet data (https://framenet.icsi.berkeley.edu/fndrupal/framenet_data) while designing or porting an application grammar.

Apart from the development of the GF code generation facility from FrameNet data and apart from a wider evaluation taking into account both more languages and more applications, future work is also to investigate the possibilities for semi-automatic multilinguality by choosing the most appropriate lexical units automatically, and by aligning lexical units (translation equivalents) among different languages.

An additional direction for future work is the application of this semantic layer to relatively unrestricted natural language – in line with the naturalist approach [1]. Angelov [19] has demonstrated the potential of the current GF Resource Grammar Library in statistical parsing of unrestricted texts (using weights extracted from a treebank). FrameNet data would provide additional means in disambiguation and would provide mapping of parse results to semantic categories. Also Barzdins [20] has addressed bridging the gap between the CNL and full natural language through use of FrameNet on the discourse level. Integration with frame semantics thus provides additional means towards semantic parsing of less controlled text.

## Acknowledgments

This work has been supported by the European Regional Development Fund under the project No. 2011/0009/2DP/2.1.1.1.0/10/APIA/VIAA/112. The authors would like to thank the reviewers for the detailed comments and constructive criticism.